\documentclass[10pt,twocolumn,letterpaper]{article}

\usepackage[pagenumbers]{cvpr} 

\usepackage{booktabs}
\usepackage{adjustbox} 
\usepackage{pifont}
\usepackage{longtable}
\usepackage{lipsum}
\usepackage{tikz}
\usepackage{amsfonts}
\usepackage{float}

\definecolor{cvprblue}{rgb}{0.21,0.49,0.74}
\usepackage[pagebackref,breaklinks,colorlinks,allcolors=cvprblue]{hyperref}

\title{MultiMAE Meets Earth Observation: Pre-training Multi-modal Multi-task Masked Autoencoders for Earth Observation Tasks}

\author{Jose Sosa\\
{\tt\small jose.sosa@uni.lu}
\and
Danila Rukhovich\\
{\tt\small danila.rukhovich@uni.lu}
\and
Anis Kacem\\
{\tt\small anis.kacem@uni.lu}
\and
Djamila Aouada\\
{\tt\small djamila.aouada@uni.lu}\\
\\
SnT, University of Luxembourg\\
}

\begin{document}
\maketitle
\begin{abstract}
Multi-modal data in Earth Observation (EO) presents a huge opportunity for improving transfer learning capabilities when pre-training deep learning models. Unlike prior work that often overlooks multi-modal EO data, recent methods have started to include it, resulting in more effective pre-training strategies. However, existing approaches commonly face challenges in effectively transferring learning to downstream tasks where the structure of available data differs from that used during pre-training. This paper addresses this limitation by exploring a more flexible multi-modal, multi-task pre-training strategy for EO data. Specifically, we adopt a Multi-modal Multi-task Masked Autoencoder (MultiMAE) that we pre-train by reconstructing diverse input modalities, including spectral, elevation, and segmentation data. The pre-trained model demonstrates robust transfer learning capabilities, outperforming state-of-the-art methods on various EO datasets for classification and segmentation tasks. Our approach exhibits significant flexibility, handling diverse input configurations without requiring modality-specific pre-trained models. Code will be available at: \textcolor{magenta}{\url{https://github.com/josesosajs/multimae-meets-eo}}
\end{abstract}
    
\section{Introduction}
\label{sec:intro}
In the Earth Observation (EO) domain, capturing and analysing remote sensing data is essential for addressing global challenges, such as resource management, natural disaster response, and environmental changes~\cite{lacoste2024geo, xiong2022earthnets, stewart2022torchgeo}. The urgent need for immediate and accurate solutions to those problems encourages the adoption of general computer vision approaches in this domain. Due to the abundant unlabelled data in EO and the inherent cost of labelling it, self-supervised learning (SSL) strategies have been preferred~\cite{cong2022satmae,noman2024rethinking,bastani2023satlaspretrain}.

\begin{figure}
    \centering
    \includegraphics[width=\linewidth]{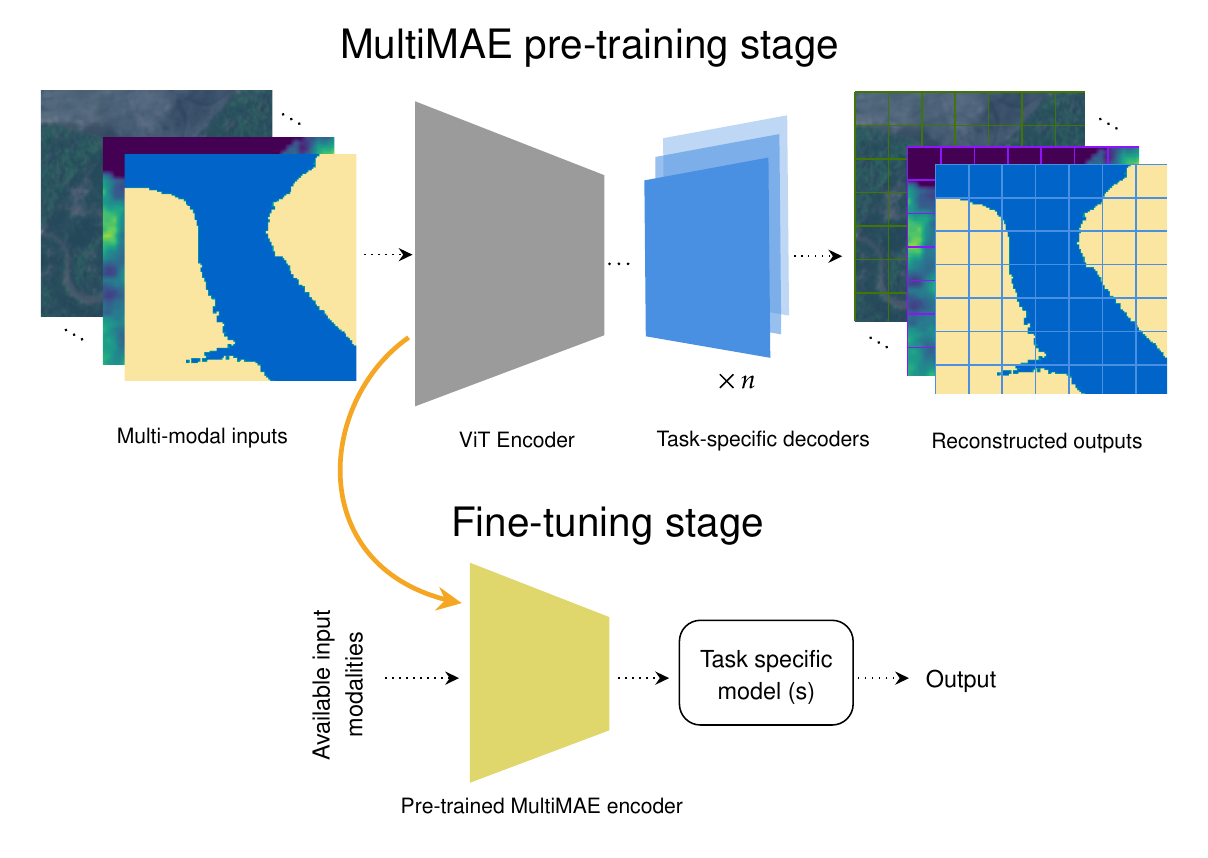}
    \caption{Pre-traning and fine-tuning stages of our MultiMAE adaptation to EO data. During pre-training MultiMAE relies on multiple input modalities. The model includes a shared ViT-based encoder and as many decoders as input modalities to support multi-tasking. When finetuning, the pre-trained encoder is coupled with the task specific model (depending on the downstream task). Note that during this stage the number of input modalities could be different from those on pre-training.}
    \label{fig:enter-label}
\end{figure}

Early attempts to apply SSL to EO data often rely on transfer learning from single-modality, general-domain datasets~\cite{deng2009imagenet}. While this approach is practical in some cases, it might not be optimal due to the
heterogeneous data and diversity of downstream tasks in EO. Thus, recent works focus on using in-domain datasets to train deep learning (DL) approaches that can serve as many downstream tasks as possible, e.g., foundation models for EO~\cite{Jakubik2023FoundationMF,bastani2023satlaspretrain,stewart2023ssl4eo, guo2024skysense, Jakubik2023FoundationMF, xiong2024neural}. Typically, those models follow a pre-training stage that involves an extensive collection of satellite imagery (commonly Sentinel-2), permitting the extraction of rich features. Then, the pre-trained model provides initialisation for downstream tasks, such as land cover classification and crop segmentation. Methods adopting this strategy achieve remarkable performance in EO tasks when using Sentinel-2 (S2) imagery for pre-training and fine-tuning~\cite{cong2022satmae, noman2024rethinking}. However, as highlighted in~\cite{sosa2024effective}, their flexibility is often compromised when the structure of fine-tuning data diverges from that of the pre-training data. 

According to previous advances in the general computer vision domain, combining multi-modality with multi-task strategies has proven effective in learning richer representations and improving performance across diverse tasks~\cite{bachmann2022multimae, lin2023multimodality}. Unfortunately, in the EO domain the multi-modal nature of this data (originated by diverse sensors) is often ignored~\cite{stewart2022torchgeo, xiong2022earthnets}. This is partly due to the lack of complete publicly available multi-modal domain-specific datasets. Nevertheless, the recent emergence of well-structured multi-modal benchmarks~\cite{nedungadi2024mmearth} constitutes a promising resource for developing multi-modal multi-task DL frameworks exclusively for the EO domain.

This paper investigates the use of multi-modal EO data during the pre-training stage of a multi-modal, multi-task Masked Autoencoder (MAE)-based architecture (MultiMAE)~\cite{bachmann2022multimae}. We argue that pre-training such model on strategically selected EO modalities can produce transferable features, improving performance and flexibility across various in-domain downstream tasks. The core architecture of our approach relies on a modified MAE~\cite{he2022masked}. It uses a Vision Transformer (ViT) encoder to jointly process different input modalities from EO data. Then, multiple modality-specific decoders reconstruct each input separately, hence supporting multi-task learning. For pre-training MultiMAE with EO data, we build modalities by splitting S2 spectral channels. Additionally, we incorporate depth information and segmentation labels from a recent multi-modal EO dataset~\cite{nedungadi2024mmearth}. Comprehensive evaluations demonstrate that our method consistently outperforms related works across multiple EO datasets for downstream classification and segmentation tasks. Our key contributions are as follows: 
\begin{itemize}
    \item We successfully adapt a multi-modal, multi-task ViT-based MAE to the EO domain. Our implementation is the first approach of its kind to explore multi-modal multi-task pre-training with data from the MMEarth dataset~\cite{nedungadi2024mmearth}. 
    \item We demonstrate that by strategically splitting S2 and treating the resulting groups as modalities for pre-training, our multi-modal multi-task ViT-based MAE offers more flexibility when fine-tuning with distinct data availability. 
    \item We conduct various experiments with many EO datasets to validate the effectiveness of pre-training a MultiMAE on multi-modal EO data. 
\end{itemize}

\section{Related Work}
\label{sec:format}

\textbf{Self-supervised learning} has been widely adopted as a pre-training approach across many computer vision tasks. It benefits transfer learning in domains where unlabelled data is abundant, like EO. Works in this direction employ different SSL strategies, including contrastive and continual learning; and most related to our work, Masked Image Modelling (MIM)~\cite{he2022masked,hondru2024masked}. Thanks to the introduction of ViTs~\cite{dosovitskiy2020image} and their suitability for MIM, these rapidly become an attractive option for SSL pre-training~\cite{he2022masked, xie2022simmim,huang2022masked}. 

In the context of EO, recent approaches such as SatMAE~\cite{cong2022satmae} and SatMAE++~\cite{noman2024rethinking} successfully adapt MAE~\cite{he2022masked} to reconstruct data that differs from standard RGB format, such as S2 imagery. However, those methods limit the fine-tuning stage to the same input structure as in pre-training. This limitation reduces flexibility for transfer learning, making it challenging to handle all those EO downstream tasks where complete S2 data is unavailable. Our approach eliminates this constraint and allows using the same pre-trained model with an arbitrary number of inputs during fine-tuning. Thereby enhancing flexibility and avoiding repeating costly pre-training processes.


\noindent \textbf{Multi-task} and \textbf{multi-modal} approaches like MultiMAE~\cite{bachmann2022multimae} are well-known for learning robust representations from unlabelled data in the general computer vision domain. However, these concepts remain relatively under-explored in the EO domain as prior methods focus solely on S2 data \cite{cong2022satmae,noman2024rethinking}. Fortunately, recent works based on MAEs start investigating multi-modal and multi-task settings, exploiting the heterogeneous data available in EO. For example, some approaches extended S2 data by incorporating text descriptions \cite{liu2024remoteclip} or geolocation information \cite{bastani2023satlaspretrain} as input modalities. Others, like \cite{nedungadi2024mmearth}, considerably increase the number of input modalities/tasks, relying on a lightweight variation of MAEs~\cite{woo2023convnext}. Subsequent works, like DOFA \cite{xiong2024neural} and CROMA \cite{fuller2024croma} explore combinations of contrastive and MIM pre-training strategies with optical and radar input modalities. However, we opt for a more straightforward approach with less complex modalities for pre-training. Our method follows a similar strategy as previous works by adopting a multi-task, multi-modal ViT-based MAE that resembles \cite{bachmann2022multimae} but focuses on exploiting simple visual modalities from EO data. Additionally, it is closely related to \cite{nedungadi2024mmearth} in terms of the pre-training dataset but differs in implementing MIM through a ViT-based MAE rather than a CNN-based architecture. This architectural choice demonstrates its effectiveness in learning transferable representations during pre-training, while providing more flexibility when fine-tuning.

\section{Approach}
\label{sec:pagestyle}

\begin{figure*}[t]
    \centering
    \includegraphics[width=\textwidth]{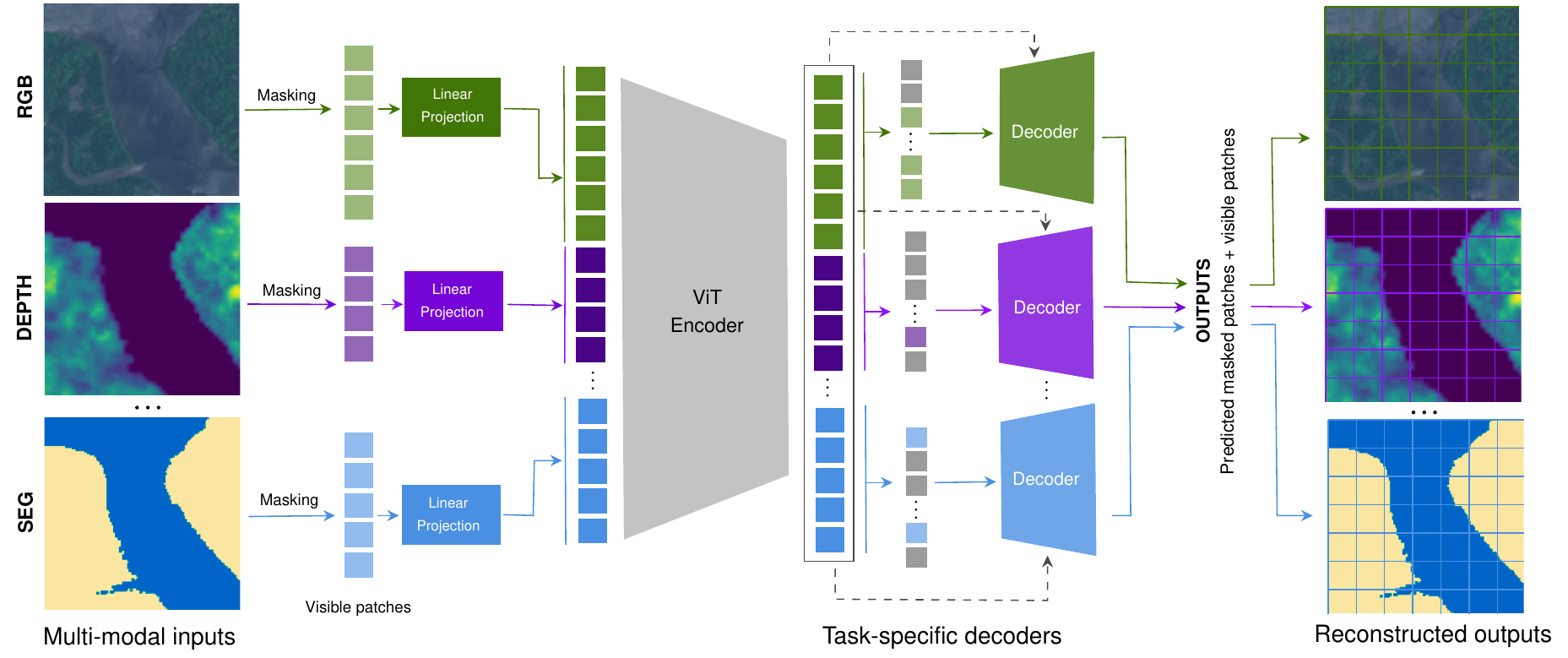}
    \caption{MultiMAE pre-training with EO data. Patches are randomly sampled from six input modalities from EO data, RGB, IRED, SIRED, EB, DEPTH, and SEG (for simplicity only three are depicted in the figure). Then, those are linearly projected and encoded via a ViT encoder. Finally, task-specific decoders reconstruct masked patches for all input modalities.}
    \label{fig:main-approach}
\end{figure*}

Our approach builds upon the Multi-modal Multi-task Masked Autoencoder (MultiMAE) architecture \cite{bachmann2022multimae}, adapting it to process visual modalities from EO data. In particular, it encodes multiple masked inputs from different visual modalities (multi-modal) through a shared ViT encoder. Then, it reconstructs each input modality separately using task-specific lightweight decoders (multi-task), as depicted in \autoref{fig:main-approach}. For pre-training our approach, we split S2 data bands to build some of the the input modalities. Additionally, we include elevation and segmentation information, enriching the shared representation. Overall, our pre-training setup considers six input modalities: four derived from S2 imagery, one from depth information, and one from segmentation labels. Following subsections detail the components of our approach.

\subsection{MultiMAE}
\label{sec:multimae}
We adapt the MultiMAE architecture~\cite{bachmann2022multimae} to work with EO data. While we follow the original implementation, we introduce key modifications to the number and structure of input modalities (tasks) and the pre-training data. 
Unlike the standard MAE~\cite{he2022masked}, which reconstructs a single input type, MultiMAE handles and reconstructs different input modalities simultaneously. The architecture follows a straightforward encoder-decoder design. It includes a shared ViT-encoder that encodes all the input modalities, and multiple decoders to reconstruct the inputs.

\noindent \textbf{Shared encoder.} Following~\cite{bachmann2022multimae}, our MultiMAE implementation relies on a ViT-based shared encoder~\cite{dosovitskiy2020image}. Each input modality is processed through dedicated patch projection layers, that convert non-masked patches into tokens. These per-modality tokens are then concatenated into a sequence and fed into the encoder. To reduce computational complexity, the encoder processes only visible tokens, omitting masked patches during encoding.

\noindent \textbf{Decoders.} We use as many decoders as input modalities to support the multi-task self-supervised reconstruction objective. Each decoder receives visible tokens corresponding to its respective modality and masked tokens. Like~\cite{he2022masked}, the decoder uses masked tokens as placeholders to reconstruct the missing patches. Additionally, the decoders take information for all the other modalities by means of a cross-attention layer, which uses the modality-specific encoded tokens as queries and the tokens for all modalities as values. The reconstruction loss is computed on masked tokens, as in~\cite{he2022masked} and~\cite{bachmann2022multimae}. Since our multi-task approach requires multiple decoders, relying on large architectures only increases complexity and pre-training requirements. To mitigate this, we rely on shallow decoders composed of a single cross-attention layer and a couple of transformer blocks, as in~\cite{he2022masked, bachmann2022multimae}.

\noindent \textbf{Masking strategy.} To keep the pre-training of MultiMAE simple and efficient, we mask out $5/6$ of the total tokens across all modalities as in \cite{bachmann2022multimae}. The visible tokens for each modality are sampled from a symmetric Dirichlet distribution, which ensures that each modality contributes to the shared representation. This provides flexibility for fine-tuning with any modality, as the sampling during pre-training is not skewed towards any particular input.


\subsection{Handling multi-spectral data} 
Unlike general domain computer vision tasks, which mostly rely on RGB images, the EO domain widely employs S2 imagery~\cite{noman2024rethinking}. Previous approaches that pre-train ViT-based MAEs on S2 data employ different strategies to handle the multiple bands~\cite{noman2024rethinking, cong2022satmae}. However, during fine-tuning, available data might not contain all the necessary bands to meet the pre-training input structure. This constraint jeopardises performance and restricts the model's applicability to a broader range of downstream tasks.

We separate the S2 bands and use the resulting groups as input modalities for pre-training MultiMAE, rather than relying on the entire set of bands as a single input modality~\cite{cong2022satmae, noman2024rethinking,fuller2024croma, xiong2024neural}. The aim of separating the S2 bands is to have modalities that align with most of the available EO datasets~\cite{lacoste2024geo}. Partly inspired by previous grouping approaches~\cite{noman2024rethinking}, we construct the input modalities from S2 data as follows:

\begin{itemize}
    \item ~\textbf{RGB.} This includes bands $B4$, $B3$, and $B2$, corresponding to red, green, and blue spectral ranges. 
    \item ~\textbf{IRED.} This comprises three \textbf{I}nfra\textbf{RED} bands, $B5$, $B6$, and $B7$, keeping a structure similar to RGB.
    \item ~\textbf{SIRED}. This includes \textbf{S}hortwave \textbf{I}nfra\textbf{RED} bands $B11$ and $B12$. 
    \item ~\textbf{EB.} This follows previous approaches by considering two \textbf{E}xtra \textbf{B}ands, $B8$ and $B8A$.
\end{itemize}

We consider the ten most widely used S2 bands, distributed across four modalities. This strategy ensures flexibility, allowing our method to handle different combinations of S2 bands during pre-training and fine-tuning. When fine-tuning, unavailable modalities are simply discarded. Thus, there is no need to replicate data to fill missing channels or train separate models for varying data types, a common challenge when working with S2 and RGB inputs~\cite{cong2022satmae,noman2024rethinking}.

\subsection{Multi-modal EO dataset}
\label{sec-modal}
In EO, multi-modal datasets are not as predominant as in other computer vision domains. Recently, the release of MMEarth~\cite{nedungadi2024mmearth} represents one of the first multi-modal collections in EO that matches ImageNet size. MMEarth includes visual and textual modalities, representing an opportunity for advancing research towards multi-modal models for EO~\cite{nedungadi2024mmearth}. We explore the use of this dataset to pre-train our MultiMAE approach. Specifically, we extend the modalities derived from S2 data-RGB, IRED, SIRED, and EB- by including elevation (DEPTH) and segmentation labels (SEG). This results in a diverse set of six modalities: \textbf{RGB}, \textbf{IRED}, \textbf{SIRED}, \textbf{EB}, \textbf{DEPTH}, and \textbf{SEG}.

\begin{table*}[t]
\centering
\begin{adjustbox}{width=\textwidth}
\begin{tabular}{llcccccccc | cccc}
\toprule
\textbf{Method} & \textbf{Backbone} & \multicolumn{2}{c}{\textbf{m-eurosat$^1$}} & \multicolumn{2}{c}{\textbf{m-brick-kiln$^2$}} & \multicolumn{2}{c}{\textbf{m-so2sat$^3$}} & \multicolumn{2}{c}{\textbf{m-bigearthnet$^4$}} & 

\multicolumn{2}{c}{\textbf{fMoW (10\%)$^5$}} & \multicolumn{2}{c}{\textbf{EuroSAT$^6$}} \\
\cmidrule(lr){3-4} \cmidrule(lr){5-6} \cmidrule(lr){7-8} \cmidrule(lr){9-10} \cmidrule(lr){11-12} \cmidrule(lr){13-14}
 &  & LP &FT& LP& FT & LP &FT & LP & FT& LP & FT & LP & FT \\
\midrule
MMEarth - S2 \cite{nedungadi2024mmearth} & ConvNeXt V2& - & - &  -& - & 33.50 & 48.70 & 36.20 & 65.10 & - &-  &-  & - \\
MMEarth - Pixel M \cite{nedungadi2024mmearth} & ConvNeXt V2 & - & - &-  & - & 38.50 & 58.20 & 36.60 & 67.50 &-  &-  &  -& - \\
MMEarth64 - Full \cite{nedungadi2024mmearth} & ConvNeXt V2 & - & - & - & - & 43.80 & 54.60 & 40.90 & 68.20 & - &  -& - & - \\
SatMAE \cite{cong2022satmae} & ViT-L &-  & - & - &-  &-  &-  &  &-  & 36.76 & 58.19 & 97.65 & 98.98 \\

\midrule
MAE \cite{he2022masked} & ViT-B & 89.00 & - &88.90& - & 50.00 & - & - &-  & - &51.79  &-  & - \\
SatMAE++ \cite{noman2024rethinking} & ViT-B & - & - & - & - & - & - & - & - & - & - & -&99.04   \\
SatMAE \cite{cong2022satmae} &ViT-B & 86.40 &- &  93.90  &-  & 46.90 & - & - & - & 35.17 & 57.2& 96.61 &  99.20\\
CROMA \cite{fuller2024croma} & ViT-B & 90.10 & - & 91.10 & - & 49.20 & - & -&-  &  \textbf{38.42} & 54.47 & \textbf{97.59} & \textbf{99.22} \\
DOFA \cite{xiong2024neural} & ViT-B & 92.20 &- & 94.70 & - & 52.10& - &-  & - &-  &-  &-  & - \\
\textbf{Ours} & ViT-B & \textbf{94.10} & \textbf{97.30} & \textbf{98.30} & \textbf{98.80} & \textbf{55.97} & \textbf{59.21} & \textbf{57.9} & \textbf{70.25} & \underline{38.06} & \textbf{59.11} & 96.20 & 99.11 \\
\bottomrule
\end{tabular}
\end{adjustbox}
\caption{Performance on EO classification tasks. Results for Linear Probing (LP) and end-to-end fine-tuning (FF) on classification tasks across four datasets from GEO-Bench~\cite{lacoste2024geo}. Additionally, the S2 version of the fMoW dataset \cite{cong2022satmae}, and the full EuroSAT dataset \cite{helber2019eurosat} to extend comparisons. All results correspond to top-1 accuracy, except for those on m-bigearthnet,
expressed in mean Average Precision (mAP).}
\label{tab:class-metrics}
\end{table*}

\section{Experiments}
\label{sec:typestyle}

\subsection{Data}
\label{sec:data}

\textbf{Pre-training data.} For the pre-training stage, we rely on some visual modalities from the MMEarth dataset~\cite{nedungadi2024mmearth}, namely Sentinel-2, Aster-DEM, and ESA worldcover. Overall, the dataset consists of 1.24 million samples distributed across different world regions. Since some of those modalities in~\cite{nedungadi2024mmearth} are redundant, we strategically select a subset of them, favouring simple and inexpensive representations for ViT MAEs. This balance between simplicity and the number of input modalities/tasks helps to keep pre-training efficient. As described in \autoref{sec-modal}, we obtain RGB, IRED, SIRED, and EB modalities from S2 data, while DEPTH modality comes from elevation information in Aster-DEM and SEG modality from land cover labels in ESA worldcover.




\noindent \textbf{Fine-tuning data.} For classification tasks, we rely on datasets from GEO-Bench~\cite{lacoste2024geo}, namely m-eurosat, m-so2sat, m-bigheartnet, and m-brick-kiln. Additionally, we explore standard datasets used in previous related works \cite{noman2024rethinking, cong2022satmae}, such as the S2 version of the fMoW dataset \cite{cong2022satmae}, and the full EuroSAT dataset \cite{helber2019eurosat}. For segmentation tasks, we use m-cashew-plantation and m-SA-crop-type, also from \cite{lacoste2024geo}. 
See appendix for more details on data.





\subsection{Multi-modal multi-task pre-training}
We pre-train our MultiMAE model following the standard MAE pre-training procedure \cite{he2022masked}. However, the multi-task setting suggests that the model reconstructs multiple inputs via task-specific decoders, which is considered in the overall loss function (more details are provided in appendix). Our implementation uses a ViT-B~\cite{dosovitskiy2020image} as encoder with a patch size of $8\times8$ pixels. We employ six different input modalities/tasks from~\cite{nedungadi2024mmearth} (\autoref{sec-modal}), following a masking strategy as described in \autoref{sec:multimae}. Note that RGB and IRED inputs have three channels; SIRED and EB contain two channels, and DEPTH and SEG comprise just one channel. Our model comprises six decoders, one for each modality/task, implemented as indicated in \autoref{sec:multimae}. We use AdamW optimiser, a cosine learning rate scheduler with a starting learning rate of 1e-6, and batch size of 128 for a single GPU. We train the model on four NVIDIA A100 GPUs for 1k epochs.

\subsection{Transfer learning on downstream EO tasks}
We evaluate the transferability of the learned representations from our pre-trained approach on downstream classification and segmentation EO tasks.

\noindent \textbf{Classification setup.} We perform linear probing (LP) and end-to-end fine-tuning (FF), using the six datasets as indicated in \autoref{sec:data}. For all classification experiments, we employ the four available S2-derived modalities as inputs, namely RGB, IRED, SIRED, and EB. We fine-tune the models with both LP and FF for a maximum of 50 epochs. The input size is $96\times96$ for each modality, while all other hyper-parameters align with~\cite{xiong2024neural} for a fair comparison. \autoref{tab:class-metrics} shows the evaluation results for LP and FF on classification tasks. Results appear in terms of the top-1 accuracy metric, except for results on m-bigearthnet dataset that are expressed in mean Average Precision (mAP).

\begin{table}[h]
\centering

\begin{adjustbox}{width=\linewidth}
\begin{tabular}{llcccc}
\toprule
\textbf{Method} & \textbf{Backbone} & \multicolumn{2}{c}{\textbf{m-SA-crop-type$^7$}} & \multicolumn{2}{c}{\textbf{m-cashew-plantation$^8$}} \\
\cmidrule(lr){3-4} \cmidrule(lr){5-6}
 &  & \textbf{FE} & \textbf{FF} & \textbf{FE} & \textbf{FF} \\
\midrule

MMEarth - S2 \cite{bachmann2022multimae} & ConvNeXt V2 & - & 36.00* & - & 79.90* \\
MMEarth - Pixel M \cite{bachmann2022multimae}& ConvNeXt V2 & - & \textbf{39.70}* & - & 81.90* \\
MMEarth64 - Full \cite{bachmann2022multimae}& ConvNeXt V2 & - & \textbf{39.70}* & - & 81.60* \\
DOFA \cite{xiong2024neural} & ViT-L & 32.10 &-  & 53.80 & - \\
\midrule
DOFA \cite{xiong2024neural} & ViT-B & 31.30 &  -& 48.30 &-  \\
\textbf{Ours}  & ViT-B & \textbf{33.79} & \underline{38.26} & \textbf{76.96} & \textbf{81.99} \\
\bottomrule
\end{tabular}
\end{adjustbox}
\caption{Performance on EO segmentation tasks. Results for frozen encoder (FE) and end-to-end fine-tuning (FF) on  m-SA-crop-type and m-cashew-plantation datasets.}
\label{tab:seg-metrics}
\end{table}

\noindent \textbf{Segmentation Setup.} For all experiments with segmentation tasks, the pre-trained encoder from MultiMAE is coupled with a segmentation head~\cite{bachmann2022multimae,liu2022convnet}. We adhere to the conventional end-to-end fine-tuning (FF) and fine-tuning with the frozen encoder (FE). 
For both settings, FF and FE, we fine-tune the model for 40 epochs. The input modalities are the same as in classification experiments. The input size is $256\times256$ for each modality. \autoref{tab:seg-metrics} shows results in terms of mIoU for two datasets from \cite{lacoste2024geo}.

\noindent \textbf{Multi-modal (S2-derived) fine-tuning.} \autoref{tab:class-metrics} and \autoref{tab:seg-metrics} present the results when fine-tuning with different datasets for classification and segmentation downstream EO tasks, respectively. In classification tasks, as illustrated by \autoref{tab:class-metrics}, our approach consistently outperforms previous methods on all the GEO-Bench datasets under both settings, LP and FF. When fine-tuning with other datasets, our approach again performs similarly or better than the current state-of-the-art. Remarkably, our method produces better results than approaches relying on larger versions of ViTs, like \cite{cong2022satmae} and those pre-trained with more modalities~\cite{fuller2024croma,xiong2024neural}. Furthermore, it surpasses all versions of~\cite{nedungadi2024mmearth} despite using nearly the same data for pre-training. As shown in \autoref{tab:seg-metrics}, when fine-tuning on segmentation EO tasks, our approach outperforms previous works with similar backbone on FF and FE setups across two datasets from \cite{lacoste2024geo}. In the case of the m-SA-crop-type dataset, \cite{nedungadi2024mmearth} achieves slightly superior performance. However, we hypothesise that this is due to their two-stage fine-tuning strategy. Altogether, results demonstrate the effectiveness of our multi-modal multi-task pre-training in learning transferable representations for EO downstream tasks.

\noindent \textbf{Single modality fine-tuning.} To demonstrate the flexibility of our approach, we perform single modality end-to-end fine-tuning using only RGB as input. \autoref{tab:eo_classeg} compares the respective metrics for all datasets on classification and segmentation tasks. Although we drastically reduce the number of fine-tuning modalities from four to one, results suggest that this change does not highly compromise performance. However, consistently higher differences in performance are observed in segmentation tasks, suggesting that more modalities could particularly benefit those tasks.

\begin{table}[ht]
\centering

\begin{adjustbox}{width=\linewidth}
\begin{tabular}{@{}lcccccc|cc@{}}
\toprule
\multicolumn{7}{c}{\textbf{Classification tasks}} &\multicolumn{2}{c}{\textbf{Seg. tasks}} \\
\midrule
\textbf{Input} /\ \textbf{Data}&
\textbf{1} & 
\textbf{2} & 
\textbf{3} & 
\textbf{4} & 
\textbf{5} & 
\textbf{6} & 
\textbf{7} & 
\textbf{8} \\ 
\midrule
\textbf{RGB}              & 96.10               & 98.50                  & 56.17            & 68.90                   & 52.55              & 98.69           & 32.40                    & 75.9                   \\
\textbf{S2} & \textbf{97.30}              & \textbf{98.80}                  & \textbf{59.21}            & \textbf{70.25}                  & \textbf{59.11}              & \textbf{99.11}           & \textbf{38.26}                   & \textbf{81.99}                  \\ 
\bottomrule
\end{tabular}
\end{adjustbox}
\caption{Performance comparison of single and multiple modality end-to-end fine-tuning. Numbers on second row correspond to datasets used, correspondence is indicated with superindexes on \autoref{tab:class-metrics} and \autoref{tab:seg-metrics}.}
\label{tab:eo_classeg}
\end{table}


\noindent \textbf{Fine-tuning with other modalities combinations.} We experiment with an extra dataset for multi-temporal crop segmentation \cite{hls-multi-temporal-crop-classification}, containing only RGB and IRED modalities. Although we ignore the temporal nature of the dataset, our approach exceeds the original method \cite{Jakubik2023FoundationMF} when fine-tuning with RGB and IRED. We also conduct single-modality fine-tuning using RGB data. Similar as in previous experiments, we observe a slight reduction in performance. In addition, we perform fine-tuning adding DEPTH modality, based on the assumption that aligning pre-training and fine-tuning modalities could boost performance \cite{bachmann2022multimae}. Originally, \cite{hls-multi-temporal-crop-classification} does not contain depth information. Thus, we opt for a similar strategy as in \cite{bachmann2022multimae} and create pseudo-labels for the dataset using an off-the-shelf method \cite{yang2024depth}. Adding the pseudo-depth to the input modalities and fine-tuning the model leads to a slight increase in performance compared to only using RGB and IRED, as \autoref{tab:prithvi-comp} depicts. Such a small increase might be due to the inaccuracies in obtaining out-of-domain pseudo-depth with \cite{yang2024depth}. 

\begin{table}[ht]
\centering
\begin{adjustbox}{width=\linewidth}
\begin{tabular}{@{}lccc@{}}
\toprule

\textbf{} & \textbf{RGB} & \textbf{RGB + IRED} & \textbf{RGB + IRED + DEPTH} \\ 
\midrule
Prithvi \cite{Jakubik2023FoundationMF} & - & 42.60 & -\\
\textbf{Ours}     & 38.44        & \textbf{43.19}              & 43.89   \\ 
\bottomrule
\end{tabular}
\end{adjustbox}
\caption{Performance comparison when fine-tuning with different modality combinations on crop segmentation~\cite{hls-multi-temporal-crop-classification}.}
\label{tab:prithvi-comp}
\end{table}

\section{Conclusions and limitations}
\label{sec:conc}

We present an approach for learning robust and transferable representations in the EO domain by pre-training a multi-modal, multi-task ViT-based Masked Autoencoder. Our method demonstrates effective transfer learning across diverse datasets for classification and segmentation EO tasks, consistently outperforming related works. Notably it exceeds approaches relying on bigger backbones or comprising more complex data modalities when pre-training. Furthermore, our implementation exhibits great flexibility during fine-tuning under different settings, including single-modality scenarios. Ultimately, our work supports the exploration of new and complete multi-modal EO datasets, which can contribute to standardising pre-training practices in this domain. While the adopted unbiased masking strategy balances modality contributions, future work could investigate other masking schemes and increase modalities (e.g., text) during pre-training to enhance generalisation.

\vspace{0.5cm}
\noindent \textbf{Acknowledgments.} We thank Rim Sleimi and Mohamed Aloulou for great discussions. This work is supported by FNR HPC BRIDGES project, with reference \textbf{HPC\_BRIDGES\slash2022\slash17978225\slash AI4CC}. Experiments were performed on Luxembourg national supercomputer MeluXina. Thanks to LuxProvide teams for their support.
{
    \small
    \bibliographystyle{ieeenat_fullname}
    \bibliography{main}
}
\clearpage
\setcounter{page}{1}
\setcounter{section}{0}
\setcounter{figure}{0}
\setcounter{table}{0}
\maketitlesupplementary


\section{Data details}

\subsection{Sentinel-2 data}
\begin{table}[H]
\centering

\begin{adjustbox}{width=\linewidth}
\begin{tabular}{l l c c}
\toprule
\textbf{Band} & \textbf{Description} & \textbf{Resolution} & \textbf{Wavelength (nm)} \\
\midrule
B1   & Ultra blue (Aerosol)        & 60  & 443  \\
B2   & Blue                       & 10  & 490  \\
B3   & Green                      & 10  & 560  \\
B4   & Red                        & 10  & 665  \\
B5   & Red edge 1 (near infrared)                & 20  & 705  \\
B6   & Red edge 2 (near infrared)                & 20  & 740  \\
B7   & Red edge 3 (near infrared)                & 20  & 783  \\
B8   & Near infrared             & 10  & 842  \\
B8A  & Red edge 4 (near infrared)             & 20  & 865  \\
B9   & Water vapor                & 60  & 940  \\
B10  & Cirrus                     & 60  & 1375 \\
B11  & Shortwave infrared 1 (SWIR)      & 20  & 1610 \\
B12  & Shortwave infrared 2 (SWIR)     & 20  & 2190 \\
\bottomrule
\end{tabular}
\end{adjustbox}
\caption{Sentinel-2 bands details. Details for each of the spectral bands composing Sentinel-2 data~\cite{noman2024rethinking}.}
\label{tab:s2-data}
\end{table}
Sentinel-2 (S2) imagery comprises 13 spectral bands extending across the visible, near-infrared (NIR), and shortwave infrared (SWIR) regions of the electromagnetic spectrum. These bands are provided at three different spatial resolutions: four bands at 10 m, six bands at 20 m, and three bands at 60 m. The detailed characteristics of these bands are summarised in \autoref{tab:s2-data}.

\subsection{Pre-training data}
\label{sub-pre-data}
\begin{figure}[ht]
    \centering
    \includegraphics[width=\linewidth]{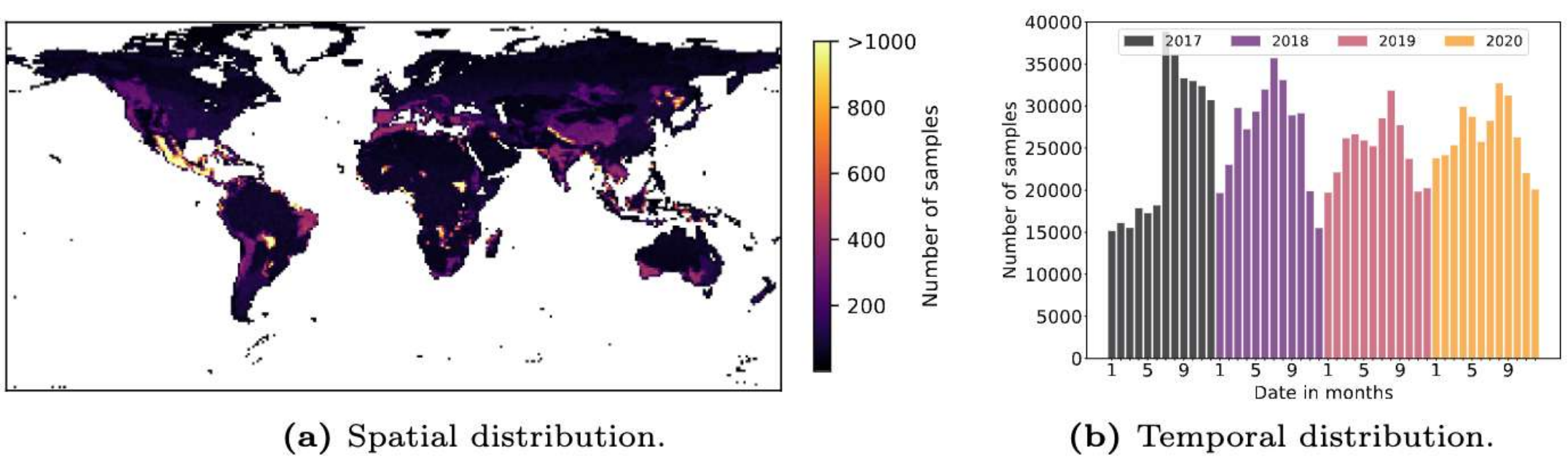}
    \caption{Spatial and temporal distribution of MMEarth dataset. Data from MMEarth spans across 4 years from multiple world regions. Multi-modal data has been collected and properly aligned using Google Earth Engine Platform. Figure taken from \cite{nedungadi2024mmearth}.}
    \label{fig:spatial-tem-dist}
\end{figure}

For the pre-training stage, we rely on the MMEarth dataset~\cite{nedungadi2024mmearth}. It represents one of the most recent and complete multi-modal large-scale collections of EO data. MMEarth matches ImageNet-1k~\cite{deng2009imagenet} size, containing 1.24 million samples. It comprises 12 aligned modalities distributed in two groups: pixel-level and image-level. The first group includes visual data, such as optical, SAR, landcover labels and elevation maps. The second group includes metadata, e.g., date, temperature information, and geolocation. \autoref{tab:pre-data} provides further details on the MMEarth dataset, while \autoref{fig:spatial-tem-dist} illustrates its spatial and temporal distribution.


\begin{table}
\centering
\begin{adjustbox}{width=\linewidth}
\begin{tabular}{lllccc}
\toprule
 \textbf{Name} & \textbf{Description}  & \textbf{Data type} & \textbf{Bands} & \textbf{Used}  \\
\hline
\textbf{Pixel-level modalities}\\
Sentinel-2 & Optical &  Continuous& 13 & $\checkmark$\\
Sentinel-1 & SAR &  Continuous& 8 & $\times$\\
Aster DEM & Elevation &  Continuous& 2 & $\checkmark$\\
ETH-GCHM & Vegetation height &  Continuous& 2 &$\times$ \\
ESA World Cover & Landcover & Categorical& 1 & $\checkmark$\\
Dynamic World & Landcover & Categorical& 1 & $\times$\\
\\
\textbf{Image-level modalities}\\
Biome & Landcover & Categorical & 1 & $\times$\\
Ecoregion & Landcover & Categorical & 1 & $\times$\\
ERA5 temperature & Climate analysis &  Continuous & 9 & $\times$\\
ERA5 precipitation & Climate analysis &  Continuous & 3 & $\times$\\
Geolocation & Latitude, Longitude &  Continuous & 4 & $\times$\\
Date & Month of the year &  Continuous & 2 & $\times$\\
\hline

\end{tabular}
\end{adjustbox}
\caption{Details of modalities from MMEarth~\cite{nedungadi2024mmearth} dataset. In this version of our approach, we strategically rely only on a subset of pixel-level (visual) modalities, as indicated by the last column of the table.}
\label{tab:pre-data}
\end{table}

\subsection{Fine-tuning data}

\begin{figure*}[t]
    \centering
    \includegraphics[width=\textwidth]{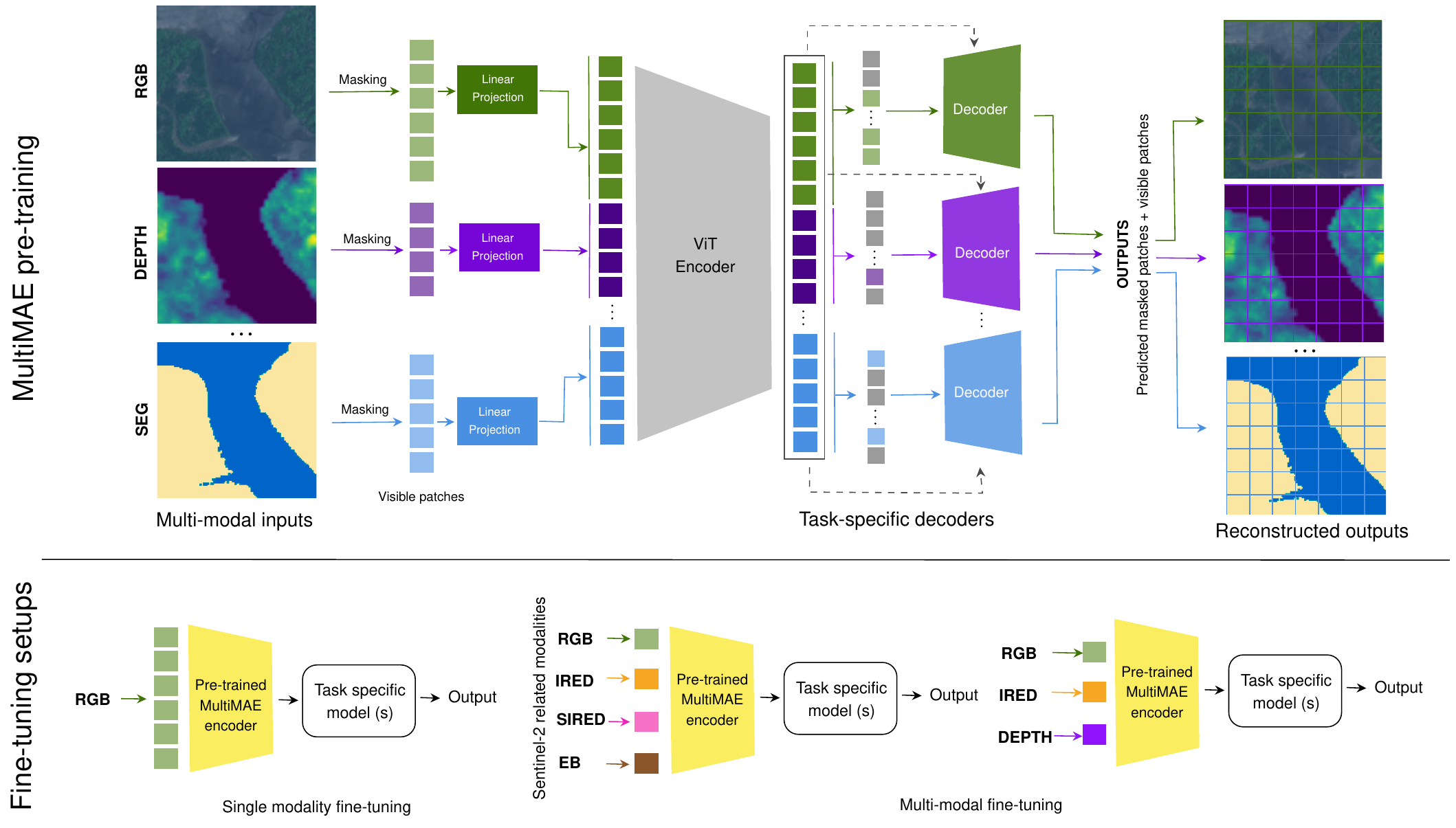}
    \caption{MultiMAE pre-training and fine-tuning with EO data. The top part of the figure illustrates the pre-training stage with six input modalities from EO data: RGB, IRED, SIRED, EB, DEPTH, and SEG (for simplicity, only three are depicted in the figure). The bottom part depicts fine-tuning setups. When fine-tuning, task-specific models are coupled with a pre-trained MultiMAE encoder. Fine-tuning occurs under multiple scenarios, e.g. single-modality or multi-modality, by varying the number of input modalities.}
    \label{fig:main-process}
\end{figure*}

For fine-tuning, we utilise mostly data from GEO-Bench~\cite{lacoste2024geo} datasets. This benchmark represents an effort to provide diverse data for fine-tuning pre-trained models on different downstream EO tasks. GEO-Bench adheres to the following design principles that make it suitable for properly evaluating the transfer learning capabilities of EO models: 

\begin{enumerate}
    \item Ease of use.
    \item Expert knowledge incorporation.
    \item Diversity of tasks.
    \item Original train, validation, and test splits.
    \item Permissive license.
\end{enumerate}

\begin{table}
\centering

\begin{adjustbox}{width=\linewidth}
\begin{tabular}{lcccc}
\toprule
\textbf{Name} & \textbf{Image Size} & \textbf{Classes} & \textbf{Train / Val / Test} & \textbf{Bands} \\
\hline
\multicolumn{5}{c}{Classification tasks} \\
m-eurosat \cite{lacoste2024geo} & $64\times64$ & 10 & 2k / 1k / 1k & 13 \\
m-brick-kiln \cite{lacoste2024geo} & $64\times64$ & 2 & 15k / 1k / 1k & 13 \\
m-so2sat \cite{lacoste2024geo} & $32\times32$ & 17 & 20k / 1k / 1k & 18 \\
m-bigearthnet \cite{lacoste2024geo} & $120\times120$ & 43 & 20k / 1k / 1k & 12 \\

EuroSAT \cite{helber2019eurosat} & $64\times64$& 10 & 16.2k / 5.4k / 5.4k   & 13\\
fMoW (10\%) \cite{cong2022satmae} & $64\times64$ & 62 & 71.3k / 85k / 85k & 13\\

\midrule
\multicolumn{5}{c}{Segmentation tasks} \\
m-SA-crop-type \cite{lacoste2024geo} & $256\times256$ & 10 & 3k / 1k / 1k & 13 \\
m-cashew-plantation \cite{lacoste2024geo} & $256\times256$ & 7 & 1.3k / 400 / 50 & 13 \\
\hline

\end{tabular}
\end{adjustbox}
\caption{EO datasets used for fine-tuning on downstream classification and segmentation tasks. Summary of datasets used for evaluating the transfer learning capabilities of our approach. Most datasets come from Geo-Bench \cite{lacoste2024geo} such as those indicated with the prefix \textit{m-}. Other standard datasets like EuroSAT~\cite{helber2019eurosat} and fMoW~\cite{cong2022satmae} are included for broader comparisons.}
\label{tab:finetuning-data}
\end{table}

Overall, \cite{lacoste2024geo} comprises multiple modified versions of standard geospatial datasets for classification and segmentation tasks. We use a subset of those datasets as shown in \autoref{tab:finetuning-data}. For fine-tuning on classification tasks, we add a couple of standard datasets used in previous related works: EuroSAT~\cite{cong2022satmae} and S2 version of fMoW~\cite{helber2019eurosat} datasets, which allows for broader comparisons. According to \cite{lacoste2024geo}, using small datasets aligns better with fine-tuning philosophy in the EO context. Thus, we reduce fMoW~\cite{cong2022satmae} and only utilise 10\% of it. Apart from this exception, all the other data collections used for fine-tuning remain unmodified.

\section{Pre-training MultiMAE}
\subsection{Pre-training objective}
\label{sub-pre}

We pre-train our approach (depicted in \autoref{fig:main-process}) using six input modalities: RGB, IRED, SIRED, EB, DEPTH, and SEG. Four of them come from Sentinel-2 data. We use all available samples in the MMEarth dataset as indicated by \autoref{sub-pre-data}. We follow a self-supervised reconstruction pre-training objective similar to standard MAEs \cite{he2022masked}. Following previous approaches \cite{he2022masked, bachmann2022multimae}, we rely on a MSE 
 (Mean Squared Error) loss on the reconstructed tokens. However, since our approach seeks to reconstruct various inputs via $N$ separate decoders $D_i$, we average the individual reconstruction losses, as indicated by \autoref{loss_eq},

\begin{equation}
\label{loss_eq}
 \mathcal{L} = \sum_{i=1}^{N} MSE(D_i(x_m,x_a), \hat{x}_m)
\end{equation}

where $x_m$ and $x_a$ correspond to the decoders inputs, i.e. modality-specific tokens and all modalities tokens, respectively, while $\hat{x}_m$ represents the ground truth tokens. In our case, $N$ is set to 6 according to the number of input modalities.

\subsection{Decoders design}

\begin{figure} [H]
    \centering
    \includegraphics[width=\linewidth]{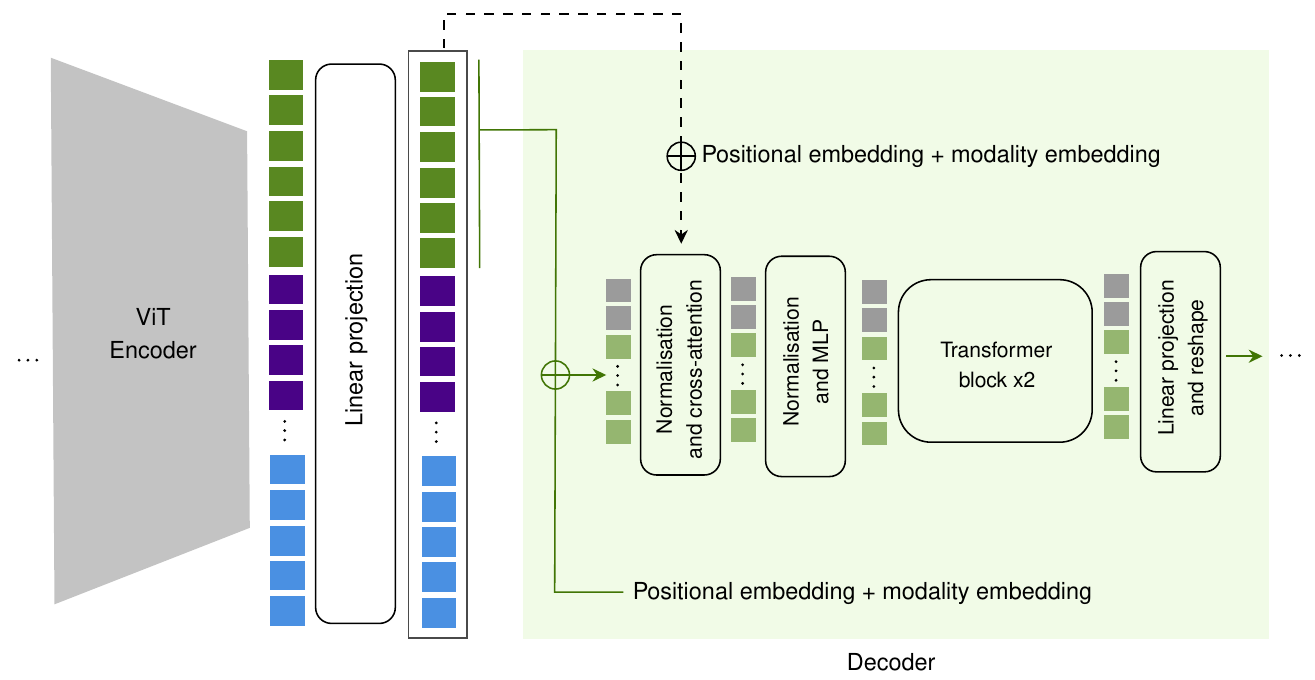}
    \caption{Decoders design. The tokens from the encoder are firstly linearly projected to match the decoder dimension. Then, modality-specific and positional embeddings are added. A cross-attention layer incorporate information from tokens of the general representation of all the modalities, which is then processed by an MLP and a couple of transformer blocks. Finally, tokens are projected and reshaped to build an image. }
    \label{fig:decoders-desing}
\end{figure}

Our decoders follow the design of those in previous works ~\cite{bachmann2022multimae, he2022masked}. Each decoder in our approach contains a linear projection layer that adapts the encoder's output to the decoder dimension. Then, after the linear projection, it adds to the decoder's inputs sine-cosine positional embeddings and the learned modality embeddings. This is further processed by a cross-attention layer, an MLP, and two transformer blocks as illustrated by \autoref{fig:decoders-desing}. Using fewer transformer blocks in the decoders makes our approach computationally efficient.

\section{Fine-tuning setups}

\begin{figure}[ht]
    \centering
    \includegraphics[width=\linewidth]{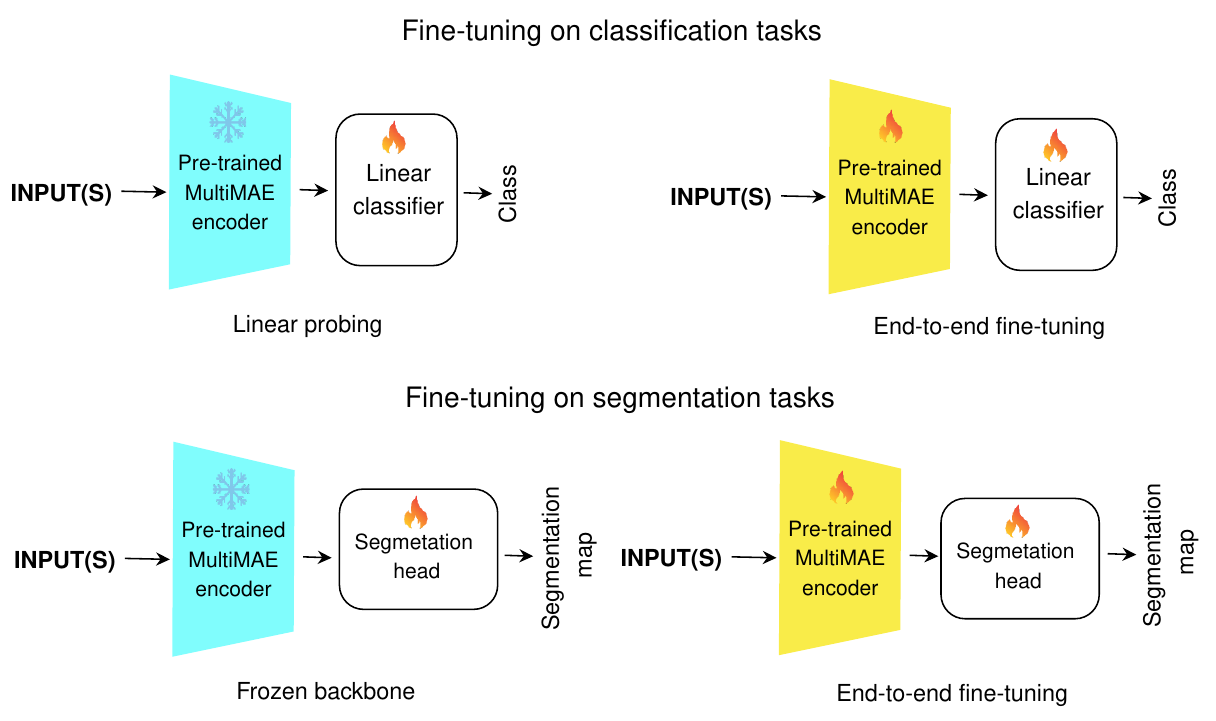}
    \caption{Fine-tuning setups for segmentation and classification EO tasks. We follow standard end-to-end fine-tuning and linear probing for classification tasks. In segmentation tasks we perform fine-tuning keeping the pre-trained encoder frozen and end-to-end fine-tuning.}
    \label{fig:finetunign-setups}
\end{figure}

For classification tasks, we couple the pre-trained MultiMAE encoder with a linear classifier. Then, we fine-tune such a model following linear probing and end-to-end fine-tuning strategies as illustrated by \autoref{fig:finetunign-setups}. During linear probing, the pre-trained encoder remains frozen, and only the parameters of the linear classifier are updated. In end-to-end fine-tuning, the pre-trained encoder and linear classifier parameters are updated. In the case of segmentation tasks, we plug a segmentation head into the pre-trained encoder. We perform fine-tuning, keeping the pre-trained encoder frozen (similar to linear probing) and standard end-to-end fine-tuning. The segmentation head consists of four ConvNeXt~\cite{liu2022convnet} blocks, which have demonstrated good alignment with ViT-based architectures \cite{bachmann2022multimae}.

\section{Qualitative results}

\subsection{Pre-training visualisations}

\begin{figure}[ht]
    \centering
    \includegraphics[width=\linewidth]{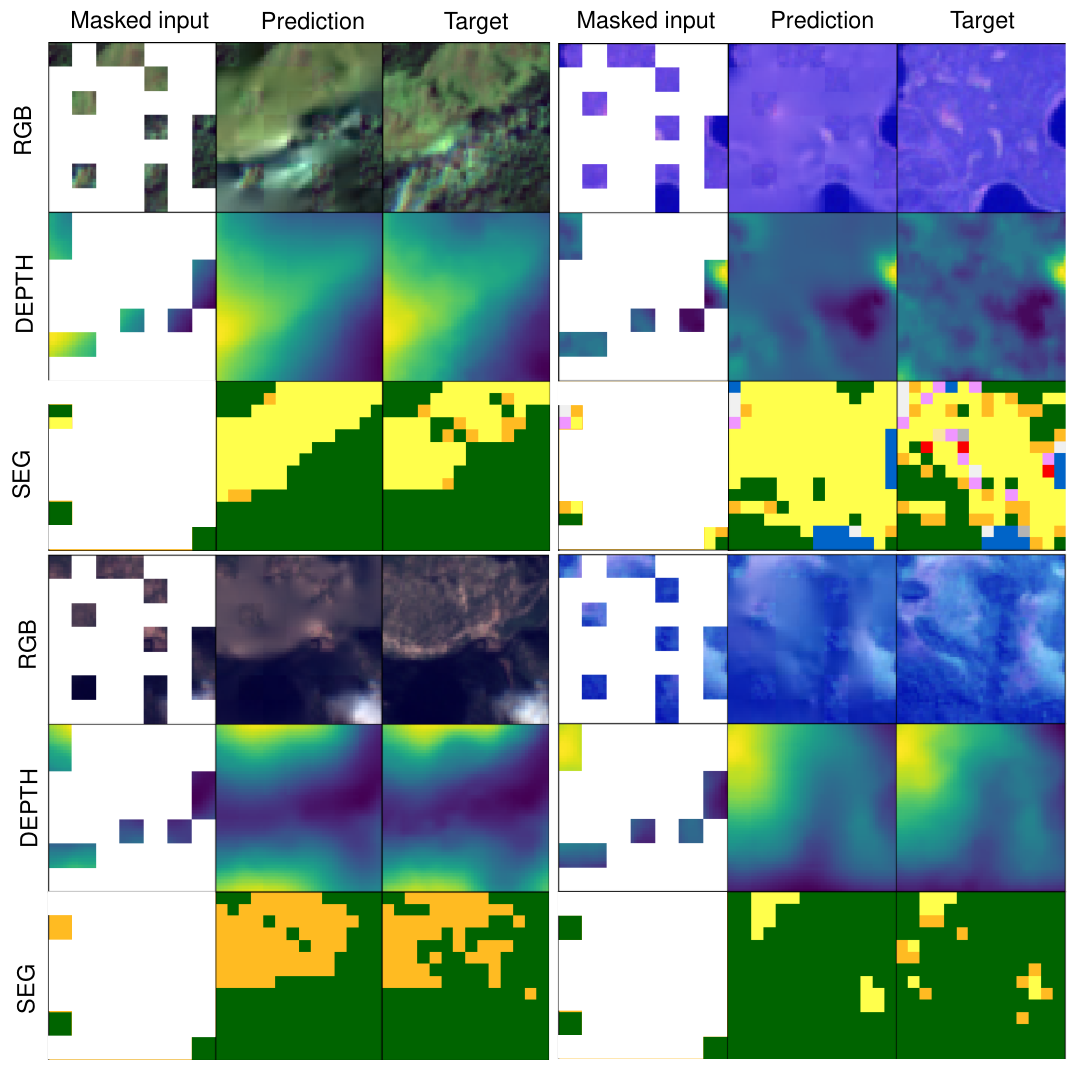}
    \caption{Visualisation of reconstructions across different input modalities. Randomly chosen reconstructions of EO input modalities after pre-training MultiMAE. The first and fourth columns depicts the masked input for RGB, DEPTH, and SEG modalities. The second and fifth columns show the reconstructed image using our approach. The third and sixth columns display the corresponding ground truth (unmasked input).}
    \label{fig:rec-vis}
\end{figure}

\autoref{fig:rec-vis} visualises randomly picked reconstructions produced by our approach. For simplicity, we only include reconstructions for RGB, DEPTH and SEG modalities within the figure. However, the pre-training stage involves the six modalities described in \autoref{sub-pre}. Note that these representations serve only illustrative purposes since they come from the training data. Based on visualisations from \autoref{fig:rec-vis}, we can notice mostly accurate reconstructions across all input modalities, which is the intended goal of the self-supervised pre-training.

\subsection{Qualitative results on segmentation tasks}

\begin{figure}[ht]
    \centering
    \includegraphics[width=\linewidth]{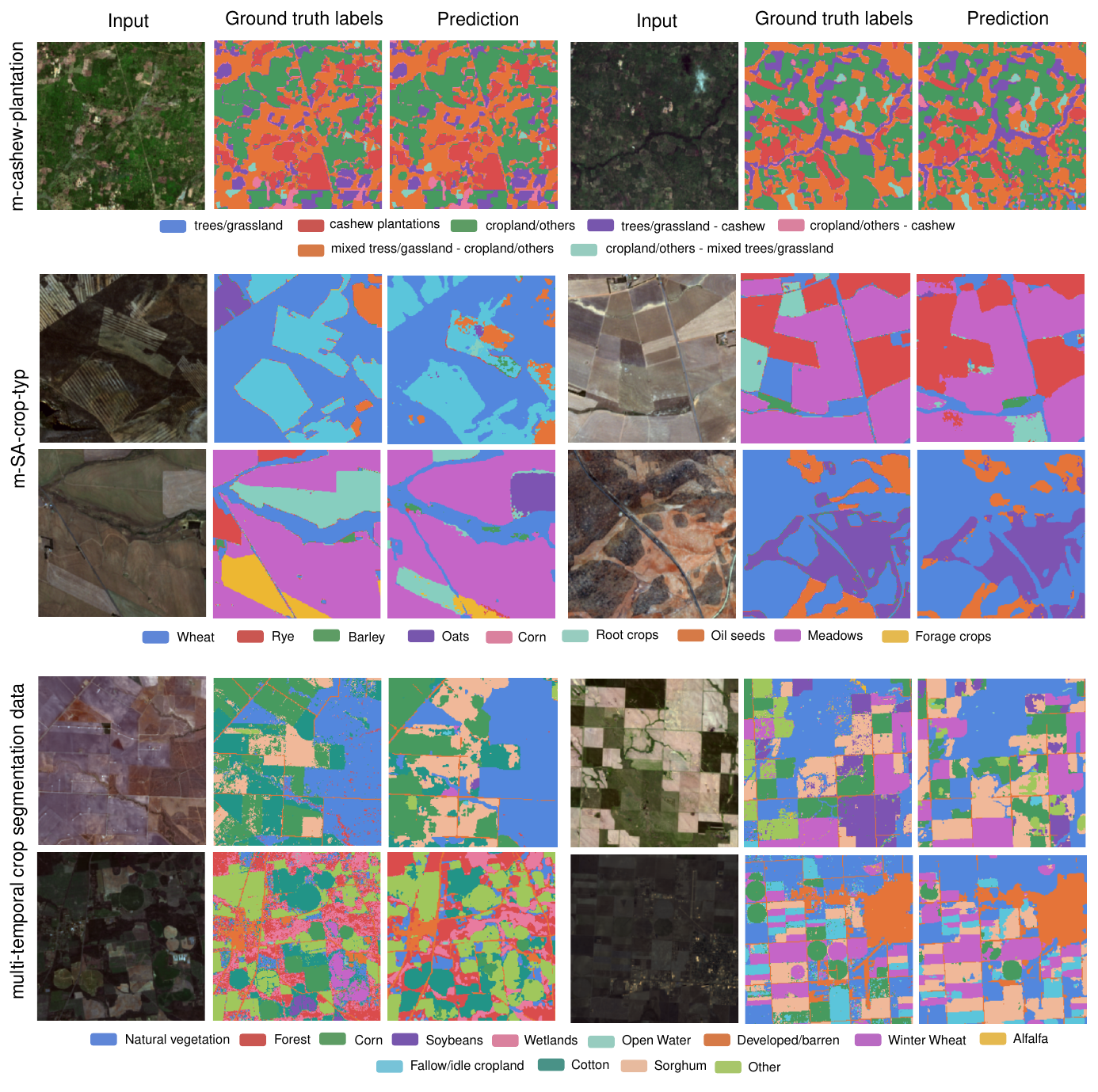}
    \caption{Visualisations for segmentation tasks. The figure visualises the predictions after fine-tuning our approach with different segmentation datasets. The first column depicts an RGB representation of the input; the second column shows the ground truth segmentation labels from the respective dataset, and the third column depicts the predicted ones by our model. Each dataset group includes a legend showing the colour code for the labels used. Labels for m-cashew-plantation correspond to specific areas useful for tracking changes in land cover. In the case of the last two datasets, segmentation labels represent crop types mostly.}
    \label{fig:seg-preds}
\end{figure}

We visualise some of the outputs after fine-tuning our approach for segmentation tasks. \autoref{fig:seg-preds} illustrates results for each of the three datasets that we use, namely m-cashew-plantation, m-SA-crop-type, and multi-temporal crop segmentation \cite{hls-multi-temporal-crop-classification}. The first column on the figure depicts a representative RGB version of the inputs. However, note that for fine-tuning, as described in the main document, S2-derived modalities were used. Specifically, the input consists of RGB, IRED, SIRED, and EB (S2-derived) modalities for m-cashew-plantation and m-SA-crop-type datasets. For the multi-temporal crop segmentation dataset, input involves RGB, IRED, and DEPTH modalities (where depth corresponds to pseudo-labels).

\end{document}